%% file: paper.tex
\RequirePackage{fix-cm}
\documentclass[smallextended]{svjour3}       
\smartqed  

\usepackage{arydshln}
\usepackage{booktabs}
\usepackage{amsmath}
\usepackage{graphicx}
\usepackage{color}
\usepackage{natbib}
\usepackage{multirow}
\usepackage{hyperref}

\newcommand{\ie}{i.e.\ }

\newcommand{\ra}{$\rightarrow$}

\newcommand{\B}[1]{\textbf{#1}}

\newcommand{\STAB}[1]{\begin{tabular}{@{}c@{}}#1\end{tabular}}

\begin{document}

\title{MSVD-Turkish: A Comprehensive Multimodal Dataset for Integrated Vision and Language Research in Turkish}
\titlerunning{MSVD-Turkish}

\author{
  Begum Citamak \and Ozan Caglayan \and Menekse Kuyu \and Erkut Erdem$^*$ \and \\
  Aykut Erdem\thanks{$^*$ Corresponding author.} \and Pranava Madhyastha \and Lucia Specia}

\institute{
    \begin{minipage}[t]{0.9\textwidth}
      Begum Citamak \and Menekse Kuyu \and Erkut Erdem
      \at Department of Computer Engineering, Hacettepe University, Ankara/Turkey \\
      \email{\{n16221821, n17132209, aykut, erkut\}\,@\,cs.hacettepe.edu.tr}\\
    \end{minipage}
     \begin{minipage}[t]{0.9\textwidth}
      Aykut Erdem
      \at Department of Computer Engineering, Ko\c{c} University, Istanbul/Turkey \\
      \email{aerdem\,@\,ku.edu.tr}\\
    \end{minipage}
    \begin{minipage}[t]{0.9\textwidth}
        Ozan Caglayan \and Pranava Madhyastha \and Lucia Specia
        \at Department of Computing, Imperial College London, UK \\
        \email{\{o.caglayan, pranava, l.specia\}\,@\,imperial.ac.uk}
    \end{minipage}
}

\authorrunning{B. Citamak et al.} 

\date{Received: date / Accepted: date}

\maketitle

\begin{abstract}
Automatic generation of video descriptions in natural language, also called {\it video captioning}, aims to understand the visual content of the video and produce a natural language sentence depicting the objects and actions in the scene. This challenging integrated vision and language problem, however, has been predominantly addressed for English. The lack of data and the linguistic properties of other languages limit the success of existing approaches for such languages. In this paper we target Turkish, a morphologically rich and agglutinative language that has very different properties compared to English. To do so, we create the first large scale video captioning dataset for this language by carefully translating the English descriptions of the videos in the MSVD (Microsoft Research Video Description Corpus) dataset into Turkish. In addition to enabling research in video captioning in Turkish, the parallel English-Turkish descriptions also enables the study of the role of video context in (multimodal) machine translation. In our experiments, we build models for both video captioning and multimodal machine translation and investigate the effect of different word segmentation approaches and different neural architectures to better address the properties of  Turkish. We hope that the MSVD-Turkish dataset and the results reported in this work will lead to better video captioning and multimodal machine translation models for Turkish and other morphology rich and agglutinative languages.
\end{abstract}

\section{Introduction}
\label{sec:introduction}
\input{sections/introduction.tex}

\section{Related Work}
\label{sec:related_work}
\input{sections/related_work.tex}

\section{MSVD-Turkish Dataset}
\label{sec:dataset}
\input{sections/dataset.tex}

\section{Modality Representations}
\label{sec:settings}
\input{sections/settings.tex}

\section{Tasks and Models}
\label{sec:approaches}
\input{sections/approaches.tex}

\section{Experimental Results}
\label{sec:results}
\input{sections/results.tex}

\section{Conclusion}
\label{sec:conclusion}
\input{sections/conclusion.tex}

\begin{acknowledgements}
This work was supported in part by TUBA GEBIP fellowship awarded to E. Erdem, and the MMVC project funded by TUBITAK and the British Council via the Newton Fund Institutional Links grant programme (grant ID 219E054 and 352343575). Lucia Specia, Pranava Madhyastha and Ozan Caglayan also received support from MultiMT (H2020 ERC Starting Grant No. 678017).

\bibliographystyle{spbasic}      
\bibliography{paper}   

\end{acknowledgements}
\end{document}

%% file: sections/introduction.tex
Recent developments in computer vision (CV) and natural language processing (NLP) have led to a surge of new problems which lie at the intersection of these two fields, creating a new area of research in general called integrated vision and language (iVL). Video captioning is one of the important problems in iVL research, which has gained significant attention in both the CV and NLP communities. It aims at understanding the visual content of a given video clip and contextually generating a natural language description of this clip.

Although considerable literature has revolved around this challenging task in recent years, all existing work is monolingual that it has mainly focused on the English language. Hence, whether or not the state-of-the-art video captioning methods can be effectively adapted to languages other than English, especially for low-resource languages,  remains an open problem. Moreover, linguistic differences between English and other languages, particularly the ones that are morphologically richer than English, introduce new challenges that need to be addressed. Before these questions can be answered, however, we require video datasets containing descriptions from languages other than English to further enable iVL research.

As a first step towards this direction, in this paper, we extend the MSVD (Microsoft Research Video Description Corpus)~\citep{msvd} dataset and introduce a new multilingual dataset that we call MSVD-Turkish which contains approximately 2k video clips and a total of 80k Turkish video descriptions. In particular, we collect these Turkish descriptions by  manually translating the original English video descriptions from MSVD into Turkish. As compared to the original English descriptions, Turkish descriptions have a larger vocabulary size and more importantly reflects the highly inflected and highly agglutinative nature of Turkish.

\begin{figure}[t]
\centering
\includegraphics[width=.9\textwidth]{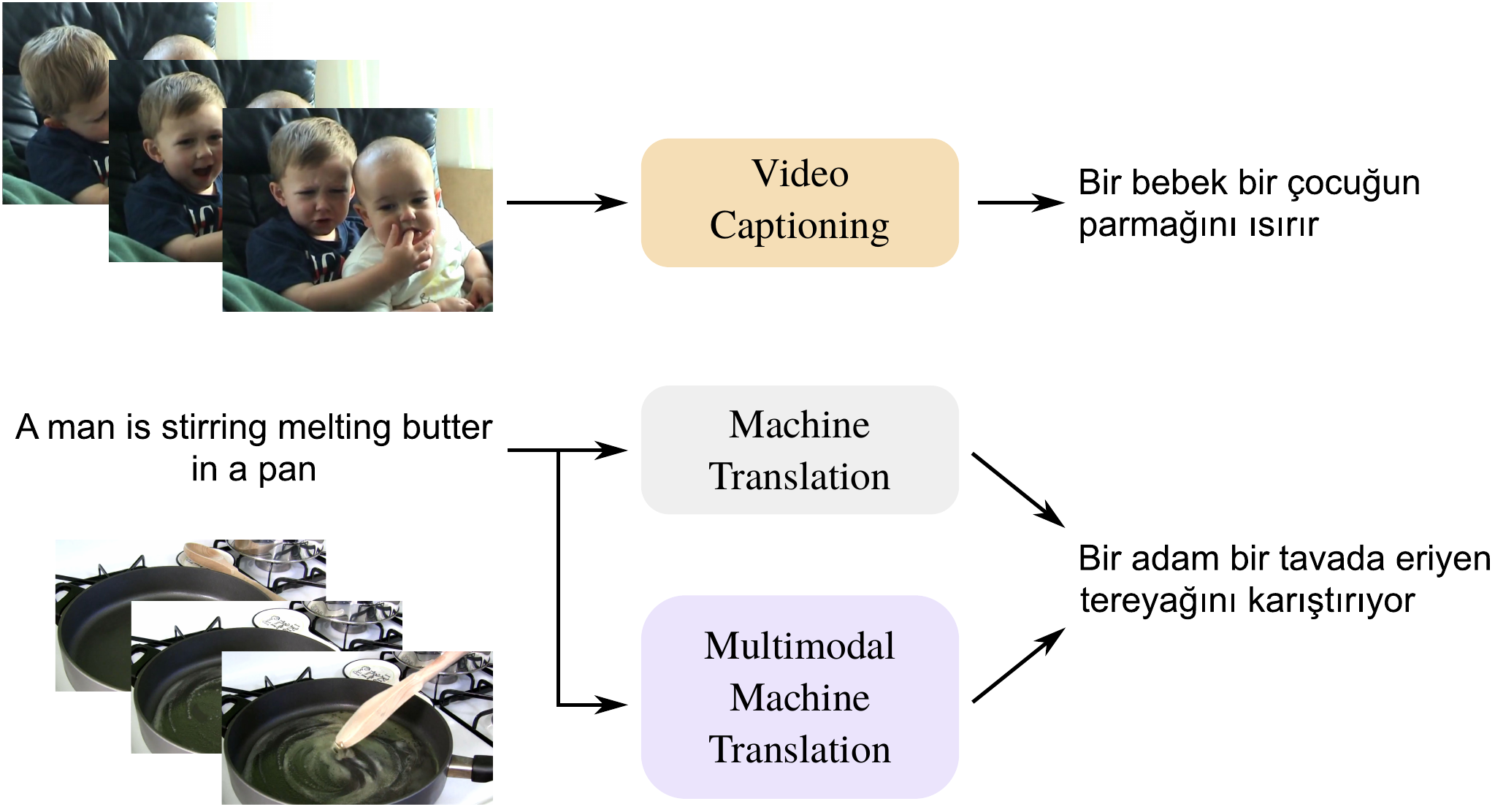}\\
\vspace*{5mm}
\caption{The depiction of the video captioning and machine translation tasks on the MSVD-Turkish Dataset.}
\label{fig:tasks}
\end{figure}

We demonstrate the multilingual, multimodal capabilities of the proposed MSVD-Turkish dataset, by exploring two distinct iVL tasks shown in Figure~\ref{fig:tasks}, namely video captioning and multimodal machine translation, but with a special focus on Turkish. As far as we are aware of, this is the first to investigate generating Turkish descriptions depicting visual content of videos. To this end, we analyse different segmentation strategies for Turkish. Additionally, we explore multimodal machine translation as the second task on MSVD-Turkish where we examine the use of supplementary visual cues within videos to potentially improve the translation quality. Our primary contributions in this paper are:
\begin{itemize}
\item To foster the research in the multilingual, multimodal language generation, we collect a new large scale dataset called MSVD-Turkish by  translating the English descriptions of the videos from the well-known MSVD dataset into Turkish.
\item We investigate the performance of several (multimodal) machine translation and video captioning models on the proposed MSVD-Turkish dataset.
\item To address the rich and agglutinative morphology of Turkish, we explore different word segmentation approaches in Turkish.
\end{itemize}

The paper is organised as follows: In Section~\ref{sec:related_work}, we briefly review the state-of-the-art in multimodal machine translation and video captioning. In Section~\ref{sec:dataset}, we introduce the MSVD-Turkish dataset, examine our data collection strategy and provide some statistics regarding the dataset. We introduce the details regarding the visual and textual representations considered in our machine translation and video captioning models in Section~\ref{sec:settings}, and describe the models themselves in Section~\ref{sec:approaches}. In Section~\ref{sec:results}, we present our experimental results and discuss our findings and finally, we provide a summary of our work and discuss possible future research directions in Section~\ref{sec:conclusion}.

%% file: sections/related_work.tex
In the following, we review the efforts towards two related tasks within the integrated vision and language research, namely multimodal machine translation and video captioning.

\subsection{Multimodal Machine Translation}
\label{sec:related_mmt}
The predominant approaches in state-of-the-art in machine translation (MT) use neural models (NMT) which consist of an encoder to map a given sentence into a latent representation, and a decoder to map this representation into a translation in the target language~\citep{Sutskever2014,cho2014gru,Bahdanau2014,transformer}. NMT models are trained with maximum likelihood estimation (MLE), \ie the training objective maximises the likelihood of source-target training pairs.

The success of such approaches has led to a rising interest in more sophisticated NMT architectures that can handle multiple input/output modalities, a framework often referred to as multimodal machine translation (MMT). MMT seeks to enhance the translation quality by taking into account visual~\citep{specia-EtAl:2016:WMT,elliott-EtAl:2017:WMT, barrault-EtAl:2018:WMT} or speech modality~\citep{sulubacak2019multimodal}. 
The prominent \textit{end-to-end} approaches to MMT with visual information can be divided into two main categories:
\begin{enumerate}
\item \textbf{Multimodal attention} extends the classical attention mechanism~\citep{Bahdanau2014} applied on top of the textual representations, with a spatial one~\citep{xu2015show} applied to convolutional feature maps. Specifically, \citet{caglayan-EtAl:2016:WMT} explore a \textit{shared attention} across the modalities while \citet{calixto-elliott-frank:2016:WMT} experiment with \textit{dedicated attention}. \citet{caglayan2016multiatt} later propose several variants where the level of parameter sharing across modality attentions is configurable.
In all these models, the outputs of attention mechanisms are simply fused together with addition or concatenation. \citet{libovicky2017attention} replace this step with another attention layer which could in theory, selectively integrate information coming from different modalities. \citet{huang-EtAl:2016:WMT} do not implement a fully multimodal attention in the decoder but enriches the source word embedding sequence with visual feature vectors, with the hope that the decoder attention will learn to pay attention to visual feature vectors when needed.

\item \textbf{Simple conditioning} makes use of non-spatial features such as the fully-connected (\texttt{FC}) layer features for VGG-style~\citep{VGG} and global average pooled features for ResNet-style~\citep{he2016resnet} networks. Specifically, a \textit{single} feature vector per image is used to condition arbitrary layers in the network, with the objective of learning \textit{grounded} textual representations. The visual conditioning is often performed through (i) initializing the hidden state of the recurrent encoders and/or decoders with the visual features~\citep{calixto-liu:2017:EMNLP2017}, (ii) multiplicative interactions of source and/or target embeddings with the visual features ~\citep{caglayan-EtAl:2017:WMT}, and (iii) the use of auxiliary training objectives such as the \textit{Imagination} architecture~\citep{elliott2017imagination} which tries to reconstruct the visual features from the textual encoder states.
\end{enumerate}

\subsection{Video Captioning}
Video captioning aims at generating a single sentence description from a short video clip summarising the actors and actions depicted in the clip. It involves unique challenges over image captioning, since it additionally requires analyzing the temporal evolvement of concepts and their relationships. The methods proposed for this task can be, in general, categorised into three classes (for a more thorough review, please refer to~\citep{Aafaq2018survey}): 

\begin{enumerate}
\item \textbf{Classical video captioning} approaches include the early works that integrate the traditional computer vision and natural language processing techniques~\citep{Kojima2002,Hakeem2004,Barbu2012,Hanckmann2012,krishnamoorthy2013generating,guadarramaiccv13,thomason-etal-2014-integrating}. These methods commonly generate description of a clip in two phases. In the first phase, they generally detect the most important objects, recognise their actions and extract object-object interactions along with the scene information. The second phase employs these extracted visual entities and some rule-based, pre-defined sentence templates to construct video descriptions. This strategy, while generates grammatically sound sentences, the sentences lack naturalness and more importantly become too constrained for open-domain videos. 

\item \textbf{Statistical video captioning} methods such as \citep{rohrbach2013translating} have been proposed to fill in this gap by additionally taking into account some statistical cues while generating a natural language description of a given input video, and thus provides more accurate and more natural depictions as compared to the classical approaches.

\item \textbf{Deep video captioning} approaches which are specifically motivated from the recent neural machine translation models. They all consider two sequential stages which are realised with an encoder-decoder architecture. The basic difference between these deep learning models and the first two groups of works lies in how they represent the visual content. While the earlier approaches employ recognition and detection methods to extract a set of word tokens, deep models represent the video in terms of a vector representation, either with a fixed or dynamic embedding. 
Deep learning based video captioning models can be categorised into further groups by their encoder-decoder structure and by their learning methodology. For instance, the most common model architecture~\citep{donahue2014long,Venugopalan2015,Yao2015} employs convolutional neural networks to extract visual content in the encoding stage and a recurrent neural network in the decoding phase to perform the video-driven sentence generation. Some other works~\citep{SrivastavaMS15,yu2015video}) extend this structure by considering recurrent neural networks in both encoding and decoding stages. The final group of studies includes reinforcement learning based video captioning models~\citep{wang2017video,yangyu2018less}. 
\end{enumerate}

%% file: sections/dataset.tex
Existing datasets for video captioning  typically contain short video clips (a few seconds in duration) and descriptions depicting the content of videos in a natural language. The early examples such as MPII Cooking \citep{mpiicooking}, YouCook~\citep{youcook}, TACoS ~\citep{regneri-etal-2013-grounding}, TACoS Multi-Level~\citep{tacosmulti}, and YouCook II~\citep{youcookii} include videos about everyday actions, which were usually collected from the video sharing sites such as YouTube by querying keywords related to cooking. On the other hand, more recent datasets such as MSVD~\citep{msvd}, M-VAD~\citep{mvad}, MPII-Movie Description (MPII-MD)~\citep{mpiimd}, TGIF~\citep{tgif}, MSR-VTT~\citep{msrvtt}, VTW \citep{vtw}, Charades~\citep{charades}, LSMDC~\citep{lsmdc}, ActyNet-Cap~\citep{actynetcap}, ANet-Entities~\citep{anetentities}, and VideoStory~\citep{videostory} are open-domain datasets. M-VAD, MPII-MD, and LSMDC datasets differ from the others in that they contain movie clips descriptions constructed by professionals for descriptive video service purposes. In recent years, with the increase in the use of social media platforms, social media has become a major source of data, and ANet-Entities~\citep{anetentities} and VideoStory ~\citep{videostory} datasets include videos shared in these mediums. Table~\ref{tab:vid_cap} summarises the characteristics of these datasets in detail.

\begin{table}[!t]
\caption{Statistics of Video Captioning Datasets}
\label{tab:vid_cap}
\renewcommand{\arraystretch}{1.3}{
\begin{tabular}{@{}l@{$\;\;$}l@{$\;\;$}l@{$\;\;$}l@{$\;\;$}l@{$\;\;$}l@{$\;\;$}l@{$\;\;$}l@{$\;\;$}l@{$\;\;$}l@{}}
\hline
\textbf{Dataset} & \textbf{Domain} & \textbf{Classes} & \textbf{Videos} & \textbf{Avg len} & \textbf{Clips} & \textbf{Sents} & \textbf{Words} & \textbf{Vocab} \\ \hline
MPII Cooking &  cooking & 65 & 44 & 600 sec & - & 5,609 & - & -  \\ 
YouCook & cooking & 6 & 88 & - & - & 2,688 & 42,457 &2,711 \\
TACoS & cooking & 26 & 127 & 360 sec & 7,206 & 18,227 & 146,771 & 28,292 \\
TACos-MLevel & cooking & 1 & 185 & 360 sec & 14,105 & 52,593 & 2,000 & - \\
MPII-MD & movie & - & 94 & 3.9 sec & 68,337 & 68,375 & 653,467 & 24,549 \\
M-VAD & movie & - & 92 & 6.2 sec & 48,986 & 55,904 & 519,933 & 17,609 \\
MSR-VTT & open & 20 & 7,180 & 20 sec & 10k & 200k & 1,856,523 & 29,316 \\
Charades & open & 157 & 9,848 & 30 sec & - & 27,847 & - & - \\
VTW & open & - & 18,100 & 90 sec & - & 44,613 & - & -  \\
YouCook II & cooking & 89 & 2,000 & 316 sec & 15.4k & 15.4k & - & 2,600 \\
ActyNet Cap & open & - & 20k & 180 sec & - & 100k & 1,348,000 & -  \\
ANet-Entities & social media & - &14,281 &180 sec &52k &- &- &- \\
VideoStory &social media &- &20k &- &123k &123k &- &- \\
\textsc{VaTeX} & open  & 600 &41.3k &- &41.3k &826k &- &-  \\
MSVD & open & 218 & 1970 &10 sec &1,970 & 80,827 & 567,874 & 12,592 \\
MSVD-Turkish & open & 218 & 1970 & 10 sec &1,970 & 80,676 & 432,250 & 18,312\\ \hline
\end{tabular}}
\end{table}

It is important to mention that all these datasets are monolingual and contain only English descriptions. The only exception is the recently proposed \textsc{VaTeX} dataset~\citep{Wang2019vatex}, which has both Chinese and English descriptions for each video clip. Even for image captioning which has been studied more extensively than video captioning, multilingual datasets are scarce. There exist only a few datasets such as (i) the TasvirEt dataset~\citep{tasvir_et} which extends the original Flickr8k dataset with two crowdsourced Turkish descriptions per image, (ii) the STAIR dataset~\citep{stair_dataset} which provides five crowd-sourced Japanese descriptions for 164,062 MSCOCO~\citep{lin2014microsoft} images. Similarly, the image-based MMT task requires a multimodal dataset with images and their (translated) descriptions in at least two languages. The well-known Multi30k dataset~\citep{Elliott2016} fulfilled this requirement by augmenting the popular image captioning dataset Flickr30k~\citep{flickr30kent}, with German, French and Czech descriptions that are direct translations of the original Flickr30k English descriptions. Multi30k is so far the only dataset which provides actual translations aligned to images, rather than independent descriptions as in TasvirEt and STAIR datasets.

In this study, we aim to contribute to this new area of research, \emph{multilingual video captioning}, by collecting a large video dataset consisting of videos and their English and Turkish descriptions. We believe that selecting Turkish will fill an important gap on the analysis of morphologically-rich and low-resource languages in the video captioning literature. We name our dataset as MSVD-Turkish, after the MSVD dataset. Since MSVD-Turkish has parallel Turkish-English sentences, it can be used not only for video captioning task but also for multimodal machine translation\footnote{We make our dataset publicly available at \url{https://hucvl.github.io/MSVD-Turkish/}}.

\begin{figure}[t]
\centering
\includegraphics[width=.99\textwidth]{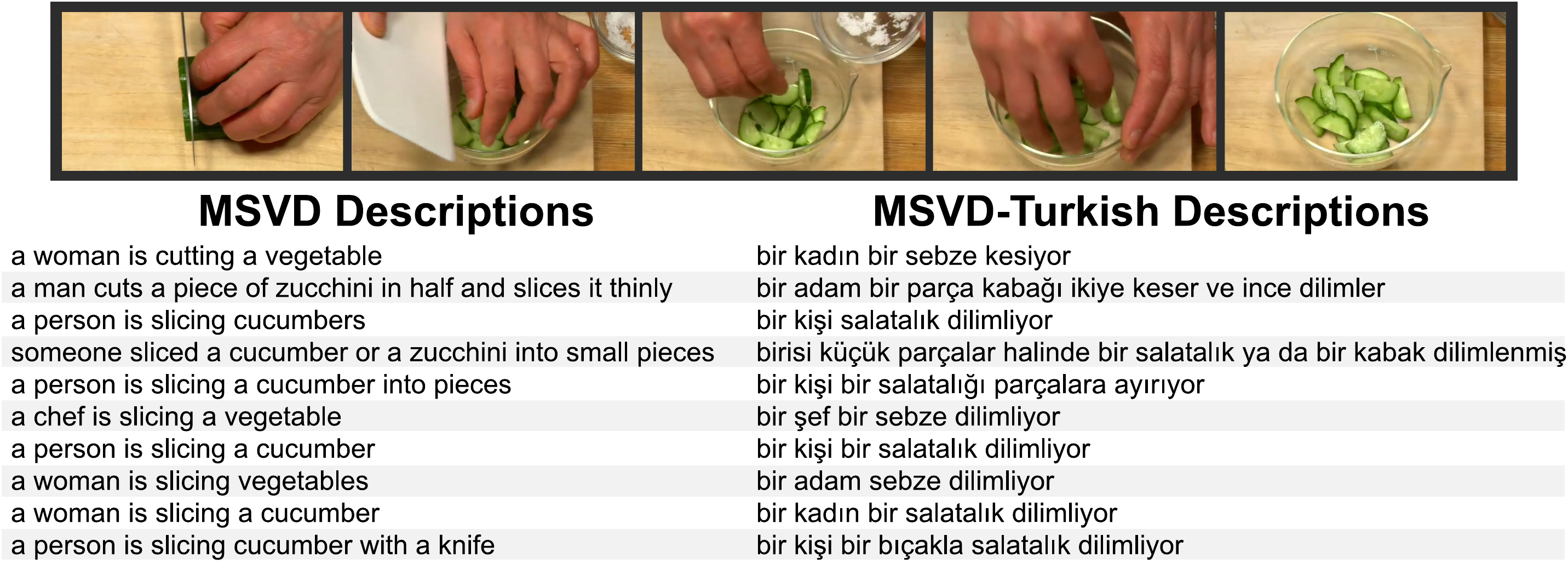}\\
\caption{A sample clip and its English captions with the corresponding Turkish translations from the MSVD-Turkish dataset.}
\label{fig:dataset_examples}
\end{figure}

In the data collection phase, we first translated the English captions into Turkish with the free Google Translate API. The main drawback of using such an automated machine translation system is that the generated translations could be of low-quality. We observed that in some of the translations, suffixes were incorrectly attached or they were completely missing. Moreover, there were some translation issues regarding ambiguous words. For this reason, we gathered a team of two M.Sc. students who are Turkish-English bilingual speakers with some experience on image captioning. They checked the automatically translated sentences for obvious errors and made the necessary corrections accordingly. During this, we noticed that some English sentences in the original MSVD dataset are very noisy -- some are not even in English. We left those sentences out and did not translate them into Turkish. Figure~\ref{fig:dataset_examples} depicts an example video clip with the original English descriptions from MSVD and their translations into Turkish as provided in the resulting MSVD-Turkish dataset.

As can be seen in Table~\ref{tab:vid_cap}, the MSVD dataset contains a total of 1,970 video clips collected from YouTube, which about 10 seconds in length, with an average of 41 descriptions per clip. The training, validation and test sets from MSVD are preserved and they contain 1200, 100 and 670 videos, respectively. The number of descriptions in MSVD-Turkish is slightly lower due to the removal of noisy captions in the original MSVD. We note that there is a substantial increase in the vocabulary size ($45\%$) for Turkish, which reflects the sparsity induced by its rich morphology. Moreover, we observe that the average number of words in the descriptions is decreased in the translations, which is expected to some extent due to the agglutinating nature of Turkish in that individual words can represent multiword expressions in English. 

\begin{figure}[!t]
\centering
\includegraphics[width=.99\textwidth]{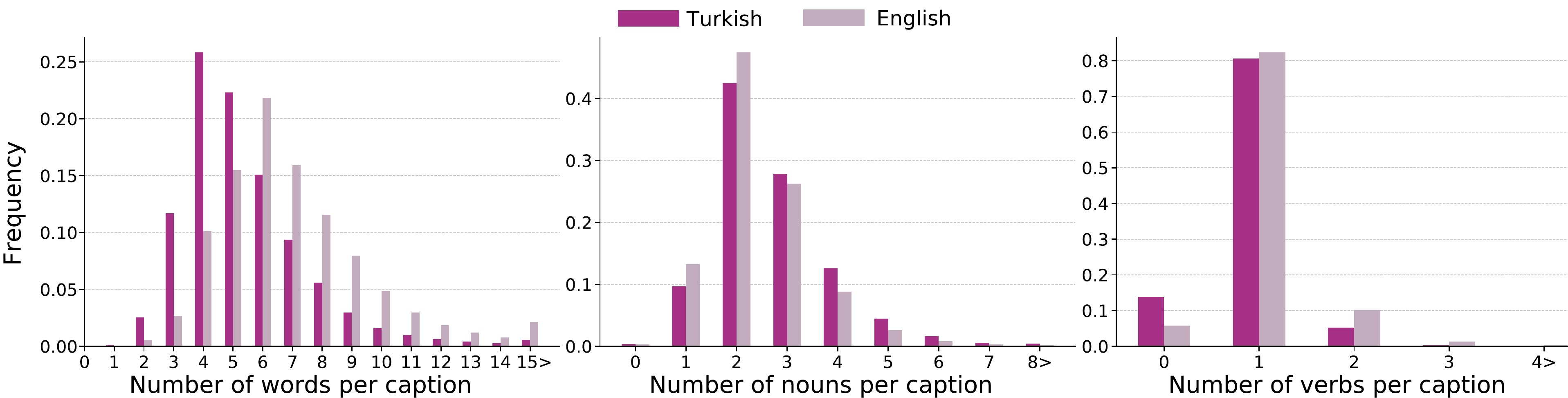}
\label{fig:dataset_statistics}
\caption{Word, noun and verb count distributions across MSVD and MSVD-Turkish datasets.}
\end{figure}

Figure~\ref{fig:dataset_statistics} presents a more comprehensive comparison between the MSVD and MSVD-Turkish datasets. Here, we utilised the \texttt{NLTK} toolkit~\citep{loper-bird-2002-nltk} for English and the \texttt{Zemberek} toolkit~\citep{zemberek} for Turkish to compute unique word, noun and verb counts in the descriptions. Linguistically, Turkish descriptions are shorter but they also contain more nouns and verbs, as compared to the case in English captions. Taking into account the differences in these distributions, we can say that the same expressions have different structures in Turkish and English.

%% file: sections/settings.tex
Both the video captioning and machine translation tasks can be cast as sequence to sequence (S2S) problems, and they involve processing of visual and/or textual information.
In the following, we briefly explain our choices for video and textual representations.

\paragraph{\textbf{Visual representations}}
We represent each video clip with a fixed-length vector by using an ImageNet pre-trained VGG16~\cite{VGG} network. In particular, we sample 30 equidistant frames from each video and feed them to VGG16 to extract 4096-dimensional \texttt{FC7} feature vectors. The final video representation that will be used by the video captioning models is this temporal feature matrix of size $30\times 4096$.
\paragraph{\textbf{Textual representations}}
In addition to the default setting of using words for both languages, we conduct several experiments using two widely-known unsupervised segmentation techniques.
The motivation behind these experiments is to understand what kind of token representation is ideal for video captioning and machine translation into Turkish, an agglutinative language with rich morphology.

The first segmentation approach makes use of the so-called ``byte-pair encoding (BPE)" algorithm, which proceeds by deconstructing sentences to characters and then works out in a bottom-up manner to recursively merge frequent byte pairs altogether \citep{sennrich2015neural}. The final ``open vocabulary" is biased towards keeping frequent words intact while splitting out rare words into frequent subwords. The main hyperparameter of the BPE algorithm determines how many merge operations will be done during the learning step, which approximately reflects the final vocabulary size. In contrast to the BPE algorithm which is deterministic, \texttt{SentencePiece} (SPM) employs a unigram language modeling approach to maximize the likelihood of a given corpus with the probability of each sentence defined as the sum of its candidate segmentations \citep{sentencepiece}. Overall, this yields a probabilistic mixture model from which it is possible to sample arbitrarily many multi-level segmentations (characters, subwords and words) for a given sentence. SPM can also be applied to non-tokenized sentences, removing the necessity of using language-dependent tokenization and detokenization pipelines.

%% file: sections/approaches.tex
In this section, we present two tasks exploring the proposed bilingual MSVD dataset where the common objective is to generate natural language sentences in Turkish. We first start by describing the sequence-to-sequence framework within the context of monolingual and multimodal machine translation. We then present the video captioning approaches that we follow, which can be considered as extensions to the S2S framework. We note that some design choices and hyperparameters are different between the neural architectures and the video captioning and MMT tasks because they were empirically selected for each architecture and task.

\subsection{Multimodal Machine Translation}
\begin{figure}[t]
\centering
\includegraphics[width=.98\textwidth]{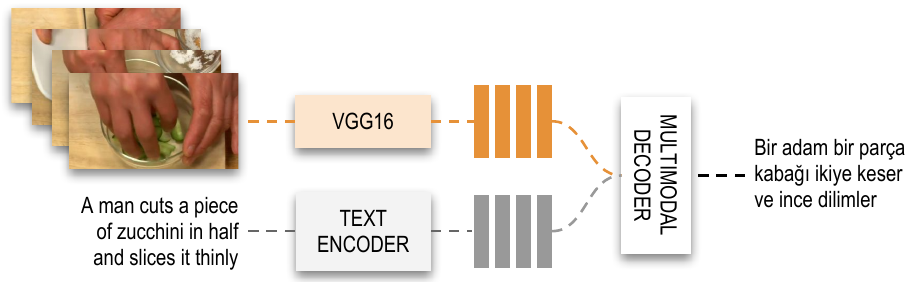}
\caption{Multimodal machine translation decoder on MSVD-Turkish.}
\label{fig:mmt_model}
\end{figure}

In what follows, we introduce our recurrent and transformer-based NMT models and their multimodal counterparts. To represent the words in the sentences, we experiment with the segmentation approaches previously mentioned in Section~\ref{sec:settings}.
To achieve multimodality, we provide the 4096-dimensional frame representations extracted from the pre-trained CNN (Section~\ref{sec:settings}) as a secondary input modality along the English sentences (Figure~\ref{fig:mmt_model}). Since videos are temporally represented by a sequence of frame features, we employ the multimodal attention approach described in Section~\ref{sec:related_mmt}. Model-specific details will be given in the respective subsections.

\subsubsection{Recurrent Model}
We follow the attentive encoder-decoder approach~\citep{Bahdanau2014} for the recurrent models. The \textit{attention mechanism} is crucial to obtain state-of-the-art results in neural machine translation (NMT). This mechanism avoids encoding the whole source sentence into a single vector as in~\cite{Sutskever2014} by looking to the latent encodings of the source sentence \textit{at each timestep} of the decoder. In other words, the decoder is conditioned on a different representation of the source sentence (namely the context $c_t$) when generating target words, rather than reusing the same fixed-size encoding vector. The model thus estimates the probability of a target token $y_t$ by conditioning on the previous target token $y_{t-1}$ and the dynamic context $c_t$ \textit{i.e.} $P(y_t | y_{t-1}, c_t)$.

\paragraph{\textbf{Architecture}} Our recurrent baseline is composed of two bidirectional GRU \citep{cho2014gru} layers in the encoder. The Turkish decoder follows the \textit{Conditional GRU} (CGRU) design~\citep{nematus} where the attention mechanism operates between two GRU layers. The input and output embeddings of the decoder are \textit{tied} for parameter efficiency~\citep{press2016using,inan2016tying}. The hidden dimensions of encoders, decoders and the attention mechanism are globally set to 320 while the source and target embeddings are 200-dimensional.
Dropout~\citep{srivastava2014dropout} is employed at three places, namely, on top of the source embeddings ($p=0.4$), encoder outputs ($p=0.5$) and before the softmax layer ($p=0.5$). We train the models using the open-source sequence-to-sequence toolkit \texttt{nmtpytorch}~\citep{nmtpy}.

\paragraph{\textbf{Multimodality}} We first experiment with the dedicated multimodal attention mechanism with concatenation~\citep{caglayan2016multiatt}. In this setup, the 4096-dimensional VGG features are projected to 640 \ie the dimension of textual encoder states. We also experiment with simple conditioning, namely by initialising the encoders and the decoders by the max-pooled VGG feature vector.

\paragraph{\textbf{Training}} We evaluate model performance after each epoch by computing the BLEU~\citep{Papineni:2002:acl} score for the validation set translations. The learning rate is \textit{halved} if the performance does not improve for two consecutive epochs. After five consecutive epochs with no improvement, the training is \textit{stopped}. At test time, the translations are decoded using the best model checkpoint, with beam size set to 12.

\subsubsection{Transformer Model}
Transformer-based NMT~\citep{transformer} are feed-forward architectures which extend the idea of attention and avoid the need for recurrent layers. This has mainly two advantages: (i) it accommodates for more layers (depth) as the gradients will flow more easily than recurrent NMTs, (ii) the removal of sequential dependence between hidden states allows parallelised training. The expressiveness of recurrence is replaced with self-attention layers which takes into account all hidden representations at a given depth. Transformer NMTs are currently state of the art, especially in large-scale NMT setups.

\paragraph{\textbf{Architecture \& Training}} Since the dataset is relatively small, we use the base Transformer model with 6 encoders and 6 decoder layers, each having 4 attention heads. The model and feed-forward dimensions are set to 256 and 1024, respectively. Label smoothing with $\epsilon = 0.1$ is applied to the cross-entropy loss. The dropout rate is fixed as 0.3. We use the open-source \texttt{fairseq} toolkit~\citep{ott2019fairseq} for training the models. The models are trained for a maximum of 50 epochs, and the test set translations are generated with beam size 12, using the checkpoint that achieved the lowest validation loss during training.

\subsection{Video Captioning}
Similar to NMT models, deep approaches to video captioning also employ encoder-decoder architectures. While the encoder module takes the individual frame representations as input and encodes them into a feature representation, the decoder generates a natural language description of the video by considering the encoded visual information. In this study, we test two popular types of network architectures, namely a recurrent one which uses the LSTM~\citep{hochreiter1997long} variant and a Transformer~\citep{transformer}. As we mentioned in Section~\ref{sec:settings}, we utilise the ImageNet pre-trained VGG16 CNN to encode the video frames, and as for the textual representation, we investigate different word segmentation strategies using SPM and BPE algorithms.

\subsubsection{Recurrent Model}
\begin{figure}[!t]
\includegraphics[width=.98\textwidth]{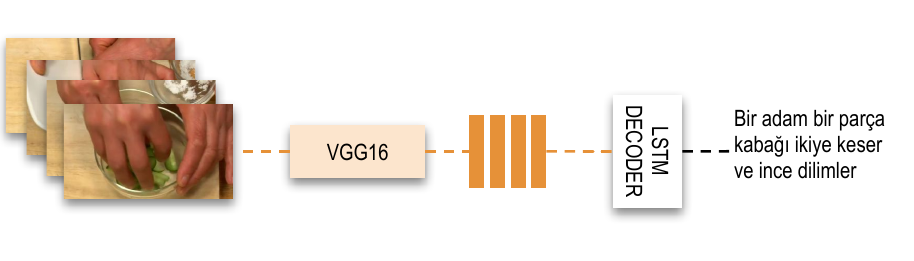}\\
\caption{Architecture of the LSTM-based video captioning model.}
\label{fig:lstm_model}
\end{figure}

\paragraph{\textbf{Architecture}} For our recurrent video captioning model, we adapt the architecture proposed by \cite{Venugopalan2015} in which the encoder and the decoder are implemented with two separate LSTM networks (Figure~\ref{fig:lstm_model}). The encoder computes a sequence of hidden states by sequentially processing the 4096-dimensional VGG16 features, extracted from the uniformly sampled video frames. The decoder module, then, takes the \textit{final hidden state} of the encoder, and accordingly outputs a sequence of tokens as the predicted video caption. There is no attention mechanism involved in this model. Both the encoder and decoder LSTM networks 
have 500 hidden units.

\paragraph{\textbf{Training}}
We use ADAM~\citep{kingma2014adam} as the optimiser and set the initial learning rate and batch size to $0.0004$ and 32, respectively. We choose the models by using early stopping. In particular, we take into account the validation loss values to decide for the checkpoint that will be used to generate Turkish descriptions at inference time.

\subsubsection{Transformer Model}
\begin{figure}[!t]
\centering
\includegraphics[width=.99\textwidth]{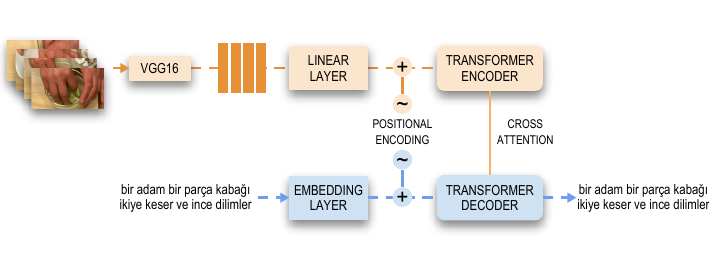}
\caption{Illustrative architecture of the Transformer-based video captioning model.}
\label{fig:transformer_model}
\end{figure}

\paragraph{\textbf{Architecture}}
Our Transformer-based video captioning model is built upon the base Transformer model~\citep{transformer}.
In the encoder, we first consider a linear transformation layer to project the extracted VGG16 features to 512. We then treat these transformed features as our visual tokens, and consider positional encodings to preserve temporal information of the frames. The decoder module is responsible for generating a description conditioned on the input video frames encoded by the encoder. Figure~\ref{fig:transformer_model} shows an illustration of our Transformer-based video captioning model.

\paragraph{\textbf{Training}} We train the models using \texttt{tensor2tensor} toolkit~\cite{tensor2tensor}. We use the base Transformer model containing 3 encoder and 3 decoder layers, each with 8 attention heads, since the dataset contains few video samples. During training, we employ the cross entropy loss with label smoothing ($\epsilon=0.1$) and a batch size of 1024. The dropout rate is fixed as 0.1. The model parameters are optimised using ADAM by setting the initial learning rate to $0.0005$. We employ approximate BLEU score for early stopping, and at test time, descriptions are obtained by using beam search with a beam size of 4.

%% file: sections/results.tex
In this section we present quantitative and qualitative results for the previously described machine translation and video captioning experiments. But before doing so, we provide the  results regarding the choice of sentence segmentation strategies based on experiments with the translation as a guidance (Section~\ref{sec:settings}).

\subsection{Machine Translation}
We first focus our attention on the word segmentation problem. For that, we use of the parallel captions from the MSVD-Turkish dataset and learn separate English and Turkish segmentation models using BPE and SPM approaches. We explore four settings where the size of the final vocabularies would be approximately 3K and 6K tokens. We train three RNN NMTs per each segmentation setup and report the mean and the standard deviation of test set BLEU scores. We also train a non-segmented word\ra word baseline for comparison.
Table~\ref{tbl:seg_results} presents the results for these experiments. We observe that in almost all cases SPM outperforms BPE, with the average gap being 0.5 BLEU in favor of SPM. Surprisingly, the best average performance 36.2 is obtained when words are used without segmentation.

\paragraph{\textbf{Pre-training segmentation models.}} These results do not favor any of the proposed segmentation approaches. We posit that this may be because of limitations of the segmentation models themselves. Therefore, we directed our attention to pre-training Turkish BPE and SPM models on a large external corpus instead of the small set of MSVD-Turkish captions. This way, we expect to learn slightly more linguistically sound segmentation models. For this purpose, we used a snapshot (2019-07-20) of the official Turkish Wikipedia dump, pre-processed it\footnote{Pre-processing consists of lowercasing, length filtering with minimum token count set to 5, punctuation removal and deduplication.} and ended up with 955K Turkish sentences.
In order to focus on the effect of segmentation in Turkish, we fix the segmentation of the English vocabulary to word units and explore only target language segmentation strategies.
Table~\ref{tbl:seg_results_enwords} presents the results for both RNN NMT and Transformer-based NMT. We see that the \texttt{SPM30K} model trained on Wikipedia performs consistently better than the others for both types of architectures. This result clearly shows the benefit of segmentation by using a large external corpus for the morphologically-rich Turkish language. We also notice that for the MSVD-Turkish dataset, the recurrent CGRU architecture slightly outperforms the Transformer models in every setting.

\begin{table}
\caption{BLEU comparison of English and Turkish segmentation choices with RNN NMT. A word\ra word baseline is also provided for comparison purposes. The segmentation models are learned on the training set of the MSVD dataset.}
\label{tbl:seg_results}
\centering
\renewcommand{\arraystretch}{1.3}{
\begin{tabular}{rccccccc}
\hline
&
\textbf{Word\ra Word} &
\textbf{3K\ra 3K} &
\textbf{3K\ra 6K} &
\textbf{6K\ra 3K} &
\textbf{6K\ra 6K} &
\textbf{Avg} \\ \hline
Words    & \B{36.2} $\pm 0.3$ & -- & -- & -- & -- & -- \\
BPE      & -- & 35.4 $\pm 0.4$  & 34.5 $\pm 0.7$   & 36.1 $\pm 0.3$  & 35.1 $\pm 0.6$    & 35.3 \\
SPM      & -- & 35.9 $\pm 0.4$  & 35.9 $\pm 0.3$   & 36.0 $\pm 0.2$  & 35.3 $\pm 0.2$    & 35.8 \\ \hline
\end{tabular}}
\end{table}

\begin{table}
\caption{BLEU comparison of Turkish Wikipedia pre-trained segmentation models for RNN and Transformer NMT models. The English vocabulary is fixed to word units and contains 9,321 words in total. The corresponding Turkish vocabulary sizes are given in parentheses.}
\label{tbl:seg_results_enwords}
\centering
\renewcommand{\arraystretch}{1.3}{
\begin{tabular}{rccc}
\hline
& 
\textbf{\ra Word (13400)} &
\textbf{\ra BPE30K (8400)} &
\textbf{\ra SPM30K (7900)} \\ \hline
Transformer    & 35.8 $\pm 0.2$ & 36.5 $\pm 0.2$ & 36.7 $\pm 0.1$   \\
RNN            & 36.2 $\pm 0.3$ & 36.8 $\pm 0.3$ & \B{37.0} $\pm 0.6$   \\ \hline
\end{tabular}}
\end{table}

\paragraph{\textbf{{Multimodal Machine Translation}}}
We now fix the choice of segmentation to words for English and to \texttt{SPM30K} for Turkish, and proceed with the multimodal machine translation results. Here, we limit the experiments to recurrent MMTs since the monomodal results did not reveal any advantages for Transformer NMT in terms of performance (Table~\ref{tbl:seg_results_enwords}).
Table~\ref{tbl:mmt} shows the results for the multimodal machine translation experiments. We observe that none of the multimodal architectures can surpass\footnote{It should be noted we reused the hyper-parameters from the NMT experiments and did not conduct a hyper-parameter search for MMTs.} the strong RNN baseline on average.
This could be because of the multimodal fusion strategy employed here and could be improved by a more sophisticated multimodal design.

\begin{table}
\caption{Multimodal machine translation results on MSVD-Turkish. The Turkish vocabulary uses \texttt{SPM30K} learned on Wikipedia and English vocabulary consists of words.}
\label{tbl:mmt}
\centering
\renewcommand{\arraystretch}{1.3}{
\begin{tabular}{lrc}
\hline
\textbf{Model} & 
\textbf{\# Params} &
\textbf{BLEU} \\ \hline
RNN                         & 8.3M  & \B{37.0} $\pm 0.6$ \\
\, + Multimodal Attention   & 11.3M & 36.5 $\pm 0.1$    \\
\, + Enc-Dec Initialisation & 9.4M  & 36.7 $\pm 0.1$    \\
\hline
\end{tabular}}
\end{table}

\subsection{Video Captioning}
In our quantitative analysis, we employ four commonly used evaluation metrics in captioning, namely BLEU~\citep{Papineni:2002:acl}, METEOR~\citep{meteor}, ROUGE~\citep{lin-2004-rouge} and CIDEr~\citep{vedantam2015cider}. The scores are computed with \texttt{coco-eval} toolkit~\citep{lin2014microsoft}. In our experiments, we train each model five times with different random seeds and report the average performances over all the runs and the corresponding standard deviations. 

As stated earlier, we experiment with both LSTM and Transformer-based architectures and analyze three different segmentation strategies, namely word, BPE and SPM level segmentations. As can be seen from the results presented in Table~\ref{tab:vid_cap_results}, Transformer-based models generate more accurate descriptions than the LSTM-based models. Moreover, switching from words to subword units extracted using SPM or BPE improves the performance in general except BLEU. Additionally, SPM and BPE based models have less number of trainable parameters than word-based models. We note that the results in Table~\ref{tab:vid_cap_results} are not directly comparable to the ones in Table~\ref{tbl:mmt} since for the machine translation experiments we consider a single reference, whereas for the video captioning experiments we consider all available references. This follows from previous work in these areas, where for MT every pair of source-reference segment is treated as an additional instance.

\begin{figure}[!t]
\centering
\includegraphics[width=0.99\textwidth]{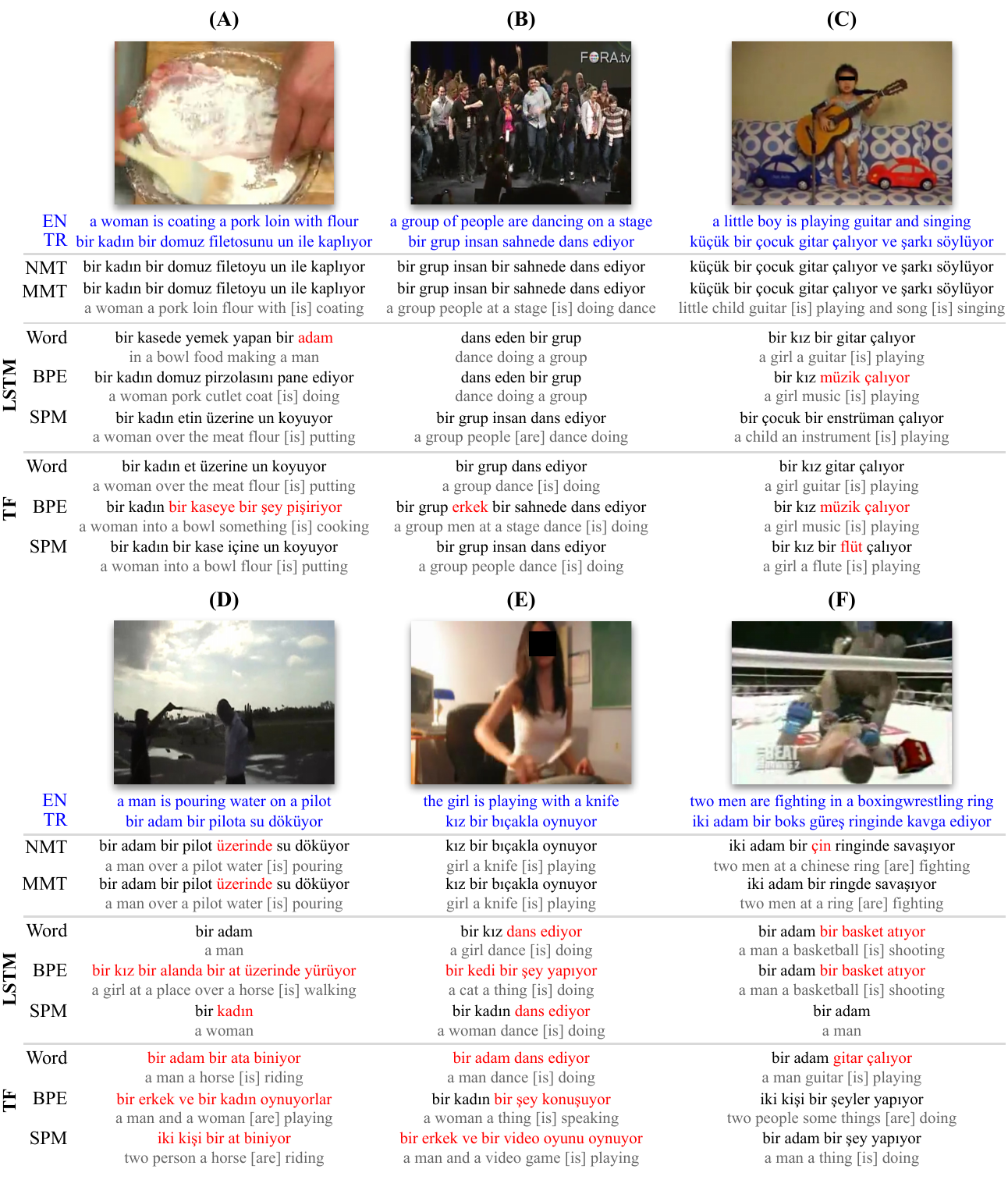}
\caption{Video captioning and machine translation results on MSVD test set: TF (Transformer) and LSTM refer to the video captioning outputs, MMT system is the \textit{enc-dec initialisation} variant from Table~\ref{tbl:mmt}. The ground-truth captions, English \textit{gloss} translations and incorrect generations are given in blue, gray and red, respectively. The examples (A) to (C) and (D) to (F) depict good and bad outputs, respectively.}
\label{fig:results}
\end{figure}

In Figure~\ref{fig:results}, we show some qualitative results of the proposed LSTM and Transformer-based Turkish captioning models and monomodal and multimodal recurrent MT models. In the top row, we show some sample results from our models where they give satisfactory translations and generations. On the other hand, we provide some corner cases in the bottom row, where the proposed models produce semantically and/or grammatically incorrect outputs. This demonstrates that there are still some open challenges and room for further research.

\begin{table}[!t]
\caption{Quantitative comparison of the LSTM and Transformer-based Turkish video captioning models in terms of BLEU, METEOR, ROUGE-L and CIDEr metrics: word, BPE, SPM-based scores and the number of trainable parameters are reported, with
\label{tab:vid_cap_results} 
\textbf{bold}-face denoting the best performance for each architecture.}
\centering
\renewcommand{\arraystretch}{1.75}{
\begin{tabular}{clccccc}
\hline
& \textbf{Vocab} & \textbf{BLEU} & \textbf{METEOR}& \textbf{ROUGE-L} & \textbf{CIDEr} & \textbf{\# Params} \\ \hline
\multirow{3}{*}{\STAB{\rotatebox[origin=c]{90}{LSTM}}} & Word  & \textbf{23.2} $\pm 1.6$ & 23.4 $\pm 0.5$& \textbf{55.3} $\pm 1.3$ & \textbf{25.4} $\pm 1.7$  & 25.2M \\
& BPE30K & 22.7 $\pm 2.6$ & \textbf{24.7} $\pm 1.4$ & 53.8 $\pm 1.8$& \textbf{25.4} $\pm 1.8$ & 17.1M \\
& SPM30K & 22.1 $\pm 1.0$& 23.9 $\pm 0.3$& 54.8  $\pm 0.4$& 24.6 $\pm 1.2$  & 16.3M \\ \hline
\multirow{3}{*}{\STAB{\rotatebox[origin=c]{90}{Transformer}}} & Word & \textbf{24.1} $\pm 0.2$ & 26.1 $\pm 0.1$ & 58.5 $\pm 0.2$ & 38.3 $\pm 0.2$ & 24.1M \\
& BPE30K & 23.8 $\pm 0.1$ & 26.7 $\pm 0.0$ & 58.4
 $\pm 0.2$ & \textbf{40.0} $\pm 0.6$ & 21.1M \\
& SPM30K & 23.9 $\pm 0.0$& \textbf{26.9} $\pm 0.0$ & \textbf{59.1}  $\pm 0.1$& 38.2 $\pm 0.1$ & 20.7M \\ \hline
\end{tabular}}
\end{table}

%% file: sections/conclusion.tex
In this paper we introduced and described a new large-scale video description dataset called MSVD-Turkish, which was constructed by carefully translating original English descriptions of MSVD dataset~\citep{msvd} to Turkish. Our dataset will allow research on novel video captioning models for Turkish, a highly inflected and  agglutinative language, as well as on multilingual video captioning approaches, including those based on translation. Additionally, as our the Turkish descriptions are direct translations of English descriptions, the dataset can be used for research in novel approaches to multimodal machine translation.

We also provided baselines using popular neural models based on recurrent neural networks and Transformer architectures. For which of these neural architectures, we analysed the use of word segmentation approaches such as BPE and SPM and demonstrated how they help both description generation and machine translation. We hope that our dataset will serve as a good resource for future efforts on multilingual, multimodal language generation. As a future work, it would be interesting to study the intrinsic annotation biases or linguistic differences between English and Turkish descriptions in MSVD and MSVD-Turkish datasets.